\documentclass{sig-alternate}
\usepackage{verbatim}

\toappear{}

\usepackage[overlay,absolute]{textpos}

\begin{document}

\begin{textblock*}{10in}(16mm, 10mm)
{\textbf{Ref:} \emph{ACM Genetic and Evolutionary Computation Conference (GECCO)}, pages 1483--1489, Montreal, Canada, July 2009.}
\end{textblock*}

\title{Simulating Human Grandmasters: Evolution 
\\and Coevolution of Evaluation Functions}

\numberofauthors{4}

\author{
\alignauthor
Eli (Omid) David\titlenote{www.elidavid.com}\\
       \affaddr{Dept.~of Computer Science}\\
       \affaddr{Bar-Ilan University}\\
       \affaddr{Ramat-Gan 52900, Israel}\\
       \email{mail@elidavid.com}\\ 
\alignauthor
H. Jaap van den Herik\\
       \affaddr{Centre for Creative Computing}\\
       \affaddr{Tilburg University}\\
       \affaddr{Tilburg, The Netherlands}\\
       \email{h.j.vdnherik@uvt.nl}\\ 
\alignauthor
Moshe Koppel\\
       \affaddr{Dept.~of Computer Science}\\
       \affaddr{Bar-Ilan University}\\
       \affaddr{Ramat-Gan 52900, Israel}\\
       \email{koppel@cs.biu.ac.il}\\ 
\and
\alignauthor
Nathan S.~Netanyahu\titlenote{Nathan Netanyahu is also affiliated with the Gonda Brain Research Center at Bar-Ilan University, and the Center for Automation Research, University of Maryland, College Park, MD 20742 (email: nathan@cfar.umd.edu)}\\
       \affaddr{Dept.~of Computer Science}\\
       \affaddr{Bar-Ilan University}\\
       \affaddr{Ramat-Gan 52900, Israel}\\
       \email{nathan@cs.biu.ac.il}\\
}

\maketitle

\begin{abstract}
This paper demonstrates the use of genetic algorithms for evolving a grandmaster-level evaluation function for a chess program. This is achieved by combining supervised and unsupervised learning. In the supervised learning phase the organisms are evolved to mimic the behavior of human grandmasters, and in the unsupervised learning phase these evolved organisms are further improved upon by means of coevolution.

While past attempts succeeded in creating a grandmaster-level program by mimicking the behavior of existing computer chess programs, this paper presents the first successful attempt at evolving a state-of-the-art evaluation function by learning only from databases of games played by humans. Our results demonstrate that the evolved program outperforms a two-time World Computer Chess Champion. 

\end{abstract}





\section{Introduction}

Despite the many advances in Machine Learning and Artificial Intelligence, there are still areas where learning mechanisms have not yielded performance comparable to humans. Computer chess is a prime example of the difficulties in such fields.

It is well-known that computer games have served as an important testbed for spawning various innovative AI techniques in domains and applications such as search, automated theorem proving, planning, and learning. In addition, the annual World Computer Chess Championship (WCCC) is arguably the longest ongoing performance evaluation of programs in computer science, which has inspired other well-known competitions in robotics, planning, and natural language understanding.

Computer chess, while being one of the most researched fields within AI, has not lent itself to the successful application of conventional learning methods, due to its enormous complexity. Hence, top chess programs still resort to manual tuning of the parameters of their evaluation function, typically through years of trial and error. The evaluation function assigns a score to a given chess position and is thus the most critical component of any chess program.

Currently, the only successful attempt reported on automatic learning of the parameter values of an evaluation function is based on ``mentor-assisted'' evolution \cite{david08a}. This approach evolves the parameter values by mimicking the evaluation function of an available chess program that serves as a ``mentor''. It essentially attempts to reverse engineer the evaluation function of this program by observing the scores it issues for a given set of positions. The approach relies heavily on the availability of the numeric evaluation score of each position, which is provided by the reference program.

In this paper, we deal successfully with a significantly more difficult problem, namely that of evolving the parameter values of an evaluation function by relying solely on the information available from games of human grandmasters, i.e., the moves played. Lacking any numeric information provided typically by a standard chess program, we combine supervised and unsupervised learning. The organisms are first evolved to mimic the behavior of these human grandmasters by observing their games, and the best evolved organisms are then further evolved through coevolution. The results show that our combined approach efficiently evolves the parameters of interest from randomly initialized values to highly tuned ones, yielding a program that outperforms a two-time World Computer Chess Champion.

In Section 2 we review past attempts at applying evolutionary techniques in computer chess. We also compare alternative learning methods to evolutionary methods, and argue why the latter are more appropriate for the task in question. Section 3 presents our new approach, including a detailed description of the framework of the GA as applied to the current problem. Section 4 provides our experimental results, and Section 5 contains concluding remarks and suggestions for future research.

\section{Learning in Computer Chess}

While the first chess programs could not pose a challenge to even a novice player, the current advanced chess programs are on par with the strongest human chess players, as the recent man vs.~machine matches clearly indicate. This improvement is largely a result of deep searches that are possible nowadays, thanks to both hardware speed and improved search techniques. While the search depth of early chess programs was limited to only a few plies, nowadays tournament-playing programs easily search more than a dozen plies in middlegame, and tens of plies in late endgame.

Despite their groundbreaking achievements, a glaring deficiency of today's top chess programs is their severe lack of a learning capability (except in most negligible ways, e.g., ``learning'' not to play an opening that resulted in a loss, etc.). In other words, despite their seemingly intelligent behavior, those top chess programs are mere brute-force (albeit efficient) searchers.

\subsection{Conventional vs. Evolutionary Learning in Computer Chess}

During more than fifty years of research in the area of computer games, many learning methods such as reinforcement learning \cite{sutton98} have been employed in simpler games. Temporal difference learning has been successfully applied in backgammon and checkers \cite{schaeffer01,tesauro92}. Although temporal difference learning has also been applied to chess \cite{baxter00}, the results showed that after three days of learning, the playing strength of the program was only 2150 Elo (see Appendix B for a description of the Elo rating system), which is a very low rating for a chess program. Wiering \cite{wiering95} provided formal arguments for the failure of these methods in more complicated games such as chess.

The issue of learning in computer chess can be seen as an optimization problem. Each program plays by conducting a search, where the root of the search tree is the current position, and the leaf nodes (at some predefined depth of the tree) are evaluated by some static evaluation function. In other words, sophisticated as the search algorithms may be, most of the knowledge of the program lies in its evaluation function. Even though automatic tuning methods, that are based mostly on reinforcement learning, have been successfully applied to simpler games such as checkers, they have had almost no impact on state-of-the-art chess engines. Currently, all top tournament-playing chess programs use hand-tuned evaluation functions, since conventional learning methods cannot cope with the enormous complexity of the problem. This is underscored by the following four points.

(1) The space to be searched is huge. It is estimated that there are about $10^{46}$ possible positions that can arise in chess \cite{chinchalkar96}. As a result, any method based on exhaustive search of the problem space is infeasible.

(2) The search space is not smooth and unimodal. The evaluation function's parameters of any top chess program are highly co-dependent. For example, in many cases increasing the values of three parameters will result in a worse performance, but if a fourth parameter is also increased, then an improved overall performance would be obtained. Since the search space is not unimodal, i.e., it does not consist of a single smooth ``hill'', any gradient-ascent algorithm such as hill climbing will perform poorly. In contrast, genetic algorithms are known to perform well in large search spaces which are not unimodal.

(3) The problem is not well understood. As will be discussed in detail in the next section, even though all top programs are hand-tuned by their programmers, finding the best value for each parameter is based mostly on educated guessing and intuition. (The fact that all top programs continue to operate in this manner attests to the lack of practical alternatives.) Had the problem been well understood, a domain-specific heuristic would have outperformed a general-purpose method such as GA.

(4) We do not require a global optimum to be found. Our goal in tuning an evaluation function is to adjust its parameters so that the overall performance of the program is enhanced.

In view of the above points it seems appropriate to employ GA for automatic tuning of the parameters of an evaluation function. Indeed, at first glance this appears like an optimization task, well suited for GA. The many parameters of the evaluation function (bonuses and penalties for each property of the position) can be encoded as a bit-string. We can randomly initialize many such ``chromosomes'', each representing one evaluation function. Thereafter, one needs to evolve the population until highly tuned ``fit'' evaluation functions emerge.

However, there is one major obstacle that hinders the above application of GA, namely the fitness function. Given a set of parameters of an evaluation (encoded as a chromosome), how should the fitness value be calculated? For many years, it seemed that the solution was to let the individuals, at each generation, play against each other a series of games, and subsequently record the score of each individual as its fitness value. (Each ``individual'' is a chess program with an appropriate evaluation function.)

The main drawback of this approach is the unacceptably large amount of time needed to evolve each generation. As a result, severe limitations were imposed on the length of the games played after each generation, and also on the size of the population involved. With a population size of 100 and a limit of 10 seconds per game, and assuming that each individual plays each other individual once in every generation, it would take 825 minutes for each generation to evolve. Specifically, reaching the 100th generation would take up to 57 days. As we see in the next section, past attempts at applying this process resulted in weak programs, which were far inferior to state-of-the-art programs.

In Section 3 we present our GA-based approach for using GA in evolving state-of-the-art chess evaluation functions. Before that, we briefly review previous work of applying evolutionary methods in computer chess.

\subsection{Previous Evolutionary Methods Applied \\to Chess}

Despite the abovementioned problems, there have been some successful applications of evolutionary techniques in computer chess, subject to some restrictions. Genetic programming was successfully employed by Hauptman and Sipper \cite{hauptman05,hauptman07} for evolving programs that can solve Mate-in-N problems and play chess endgames.

Kendall and Whitwell \cite{kendall01} used evolutionary algorithms for tuning the parameters of an evaluation function. Their approach had limited success, due to the very large number of games required (as previously discussed), and the small number of parameters used in their evaluation function. Their evolved program managed to compete with strong programs only if their search depth (lookahead) was severely limited.

Similarly, Aksenov \cite{aksenov04} employed genetic algorithms for evolving the parameters of an evaluation function, using games between the organisms for determining their fitness. Again, since this method required a very large amount of games, it evolved only a few parameters of the evaluation function with limited success. Tunstall-Pedoe \cite{tunstall91} also suggested a similar approach, without providing an implementation.

Gross \emph{et al.}~\cite{gross02} combined genetic programming and evolution strategies to improve the efficiency of a given search algorithm using a distributed computing environment on the Internet.

David, Koppel, and Netanyahu \cite{david08a} used ``mentor-assisted'' evolution for reverse engineering the evaluation function of a reference chess program (the ``mentor''), thereby evolving a new comparable evaluation function. Their approach takes advantage of the evaluation score of each position considered (during the training phase), that is provided by the reference program. In fact, this numeric information is key to simulating the program's evaluation function. In other words, notwithstanding the high-level performance of the evolved program, the learning process is heavily dependent on the availability of the above information.

In this paper, we combine supervised evolution and unsupervised coevolution for evolving the parameter values of the evaluation function to simulate the moves of a human grandmaster, without relying on the availability of evaluation scores of some computer chess program. As will be demonstrated, the evolved program is on par with today's strongest chess programs. 

\section{Evolution and Coevolution of Evaluation Functions}

Encoding the parameters of an evaluation function as a chromosome is a straightforward task, and the main impediment for evolving evaluation functions is the difficulty of applying a fitness function (a numerical value representing how well the organism performs). However, as previously noted, establishing the fitness evaluation by means of playing numerous games between the organisms in each generation (i.e., single-population coevolution) is not practical.

As mentioned earlier, the fitness value in mentor-assisted evolution is issued as follows. Both an organism and a grandmaster-level chess program are run on a given set of positions; for each position the difference between the evaluation score computed by the organism and that computed by the reference program is recorded. The fitness value is taken to be inversely proportional to this difference.

In contrast, no evaluation scores of any chess program are assumed available in this paper, and we only make use of (widely available) databases of games of human grandmasters. The task of evolution, in this case, is thus significantly more difficult than that based on an existing chess program, as the only information available here consists of the actual moves played in the positions considered.

The evaluation function is evolved by learning from grandmasters according to the steps shown in Figure \ref{fig:mentor}.

\begin{figure}[htbp]
\begin{center}
\line(1,0){240}

\begin{enumerate}
\item Select a list of positions from games of human grandmasters. For each position store the move played.
\item For each position, let the organism perform a 1-ply search and store the move selected by the organism.
\item Compare the move suggested by the organism with the actual move played by the grandmaster. The fitness of the organism will be the total number of ``correct'' moves selected (where the organism's move is the same as the grandmaster's move).
\end{enumerate}

\line(1,0){240}
\caption{Fitness function for supervised evolution of evaluation functions.}
\label{fig:mentor}
\end{center}
\end{figure}

Although performing a search for each position appears to be a costly process, in fact it consumes little time. Conducting a 1-ply search amounts to less than a millisecond for a typical chess program on an average machine, and so one thousand positions can be processed in one second. This allows us to use a large set of positions for the training set.

The abovementioned process, which will be discussed below in greater detail, results in a grandmaster-level evaluation function (see next section). Due to the random initialization of the chromosomes, each time the above process is applied, a different ``best evolved organism'' is obtained. Comparing the best evolved organisms from different runs, we observe that even though they are of similar playing strength, their evolved parameter values differ, and so does their playing style.

After running the supervised evolution process a number of times, we obtain several evolved organisms. Each organism is the best evolved organism from one complete run of the evolutionary process. We next use a coevolution phase for further improving upon the obtained organisms. During this single-population coevolution phase the evolved organisms play against each other, and the fitness function applied is based on their relative performance. Completing this phase for a predetermined number of generations, the best evolved organism is selected as the best overall organism. According to the results in the next section, this ``best of best'' organism improves upon the organisms evolved from the supervised phase. As noted before, previous attempts at applying coevolution have failed to produce grandmaster-level evaluation functions. The difference here is that the population size is small (we used 10), and the initial organisms are already well tuned (in contrast to randomly initialized).

In the following subsections, we describe in detail the chess program, the implementation of the supervised and unsupervised evolution, and the GA parameters used.

\subsection{The Chess Program and the Evaluation Function}

Our chess program uses \textsc{NegaScout}/PVS \cite{campbell83,reinfeld83} search, in conjunction with standard enhancements such as null-move pruning \cite{beal89,david08b,donninger93}, internal iterative deepening \cite{anantharaman91,scott69}, dynamic move ordering (history + killer heuristic) \cite{akl77,gillogly72,schaeffer83,schaeffer89}, multi-cut pruning \cite{bjornsson98,bjornsson01}, selective extensions \cite{anantharaman91,beal95} (consisting of check, one-reply, mate-threat, recapture, and passed pawn extensions), transposition table \cite{nelson85,slate77}, and futility pruning near leaf nodes \cite{heinz98a}.

The evaluation function of the program (which we are interested in tuning automatically) consists of 35 parameters. Even though this is a small number of parameters in comparison to other top programs, the set of parameters used does cover all important aspects of a position, e.g., material, piece mobility and centricity, pawn structure, and king safety. 

The parameters of the evaluation function are represented as a binary bit-string (chromosome size: 224 bits), initialized randomly. The value of a pawn is set to a fixed value of 100, which serves as a reference for all other parameter values. Except for the four parameters representing the material values of the pieces, all the other parameters are assigned a fixed length of 6 bits per parameter. Obviously, there are many parameters for which 3 or 4 bits suffice. However, allocating a fixed length of 6 bits to all parameters ensures that \emph{a priori} knowledge does not bias the algorithm in any way.

Note that the program's evaluation function is randomly initialized, i.e., other than knowing the rules of the game, the program has essentially no game skills at all at this point.

\subsection{Supervised Evolution using Human\\ Grandmaster Games}

As indicated, our goal is to evolve the parameters of a program's evaluation function, so as to simulate the moves played by grandmasters for a given set of positions.

For our experiments, we use a database of 10,000 games by grandmasters of rating above 2600 Elo, and randomly pick one position from each game. We pick winning positions only, i.e., positions where the side to move ultimately won the game (e.g., if it is white's turn to move, the game was won eventually by white). Of these 10,000 positions, we select 5,000 positions for training and 5,000 for testing.

In each generation, for each organism we translate its chromosome bit-string to a corresponding evaluation function. For each of the $N$ test positions (in our case, $N=5,000$), the program performs a 1-ply search using the decoded evaluation function, and the best move returned from the search is compared to that of the grandmaster in the actual game. The move is deemed ``correct'' if it is the same as the move played by the grandmaster, and ``incorrect'' otherwise. The fitness of the organism is calculated as the square of the total number of correct moves.

Note, unlike the mentor-assisted approach for mimicking an existing chess program, which provides numeric values for each position, here we only have 1-bit of information for each processed position (correct/incorrect). This underscores why relying on human games is much more difficult than using computers as mentors.

Other than the special fitness function described above, we use a standard GA implementation with Gray coded chromosomes, fitness-proportional selection, uniform crossover, and elitism (the best organism is copied to the next generation). The following parameters are used:
\\
\\
population size = 100\\
crossover rate = 0.75\\
mutation rate = 0.005\\
number of generations = 200

\subsection{Coevolution of the Best Evolved\\ Organisms}

Rerunning the supervised evolution ten times, we obtain ten ``best organisms'' corresponding to the various runs. The evaluation functions of these evolved organisms do not have the same evolved parameter values, since each run produces different results (due to the random initialization). Although the ten programs are of similar playing strength, their playing style is somewhat different. At any rate, the above ten best organisms are used for the coevolution phase described below. Note that selecting, instead, the top ten evolved organisms from one of the supervised runs is not desirable, as it could result in ``inbreeding'', in the sense that the parameter values of these organisms tend to be fairly similar.

Consider, on the other hand, generating multiple evolved organisms using different training sets for each run. Specifically, for each run we might pick games of a specific grandmaster, in the hope of obtaining organisms that mimic the individual styles of the various grandmasters. Preliminary tests suggest, however, that this variant provides no additional insight or improvement. Apparently, the 1-ply searches enable mimicking only a ``generic'' grandmaster style, rather than the style of a specific player. 

In the coevolution phase, the ten best organisms selected serve as the initial population, which is then coevolved over 50 generations. In each generation, each organism plays four games against each other organism (to obtain a more reliable result). At the end of each generation, rank-based selection is applied for selecting the organisms for breeding. Elitism is used here as well, which ensures that the best organism survives for the next generation. This is especially critical in light of the small population size. Other GA parameters remain unchanged, that is, uniform crossover with crossover rate of 0.75 and mutation rate of 0.005.

In the following section we present our experimental results, both in terms of the learning efficiency and the performance gain of the best evolved individual.

\section{Experimental Results}

We now present the results of running the evolutionary process described in the previous section. We also provide the results of several experiments that measure the strength of the evolved program in comparison to \textsc{Crafty}, a former two-time World Computer Chess Champion that is commonly used as a baseline for testing chess programs.

\subsection{Results of Supervised Evolution}

Running the evolution for 200 generations, the average number of solved positions (i.e., the number of correct moves found) increases until stabilizing at around 1,500 (out of 5,000), which corresponds to 30\% of the positions. The best organism at generation 200 solves 1,621 positions, which corresponds to 32.4\% of the positions. Due to the use of elitism, the number of solved positions for the best organism is monotonically increasing, since the best organism is preserved. The entire 200-generation evolution took approximately 2 hours on our machine (see Appendix A).

At first glance, a solution rate of 32\% might not seem too high. However, considering that the evolved organism selects successfully the ``correct'' move in one out of three cases, by applying merely a 1-ply search (as opposed to the careful analysis of a position by the grandmaster), this is quite satisfactory.

With the completion of the learning phase, we used the additional 5,000 positions set aside for testing. We let our best evolved organism perform a 1-ply search on each of these positions. The number of correctly solved positions was 1538 (30.7\%). This indicates that the first 5,000 positions used for training cover most types of positions that can arise, as the success rate for the testing set is close to the success rate for the training set.

To measure the performance of the best evolved organism after the supervised evolution phase (we call this program \textsc{Evol*}), we conducted a series of matches against the chess program \textsc{Crafty} \cite{hyatt90}. \textsc{Crafty} has successfully participated in numerous World Computer Chess Championships (WCCC), and is a direct descendent of \textsc{Cray Blitz}, the WCCC winner of 1983 and 1986. It is frequently used in the literature as a standard reference. 

Table~\ref{tab:results1} provides the results of 500 games between \textsc{Evol*} and \textsc{Crafty}. The results show that the evolved organism (\textsc{Evol*}) is on par with \textsc{Crafty}, clearly demonstrating that the supervised evolution has succeeded in evolving a grandmaster-level evaluation function by purely mimicking grandmaster moves.

\begin{table}[htbp]

\begin{center}
\begin{tabular}{|l||c|c|c|}
\hline
Match & Result & W\% & RD\\
\hline
\hline
\textsc{Evol*} - \textsc{Crafty} & 254.5 - 245.5 & 50.9\% & $+6$\\
\hline
\end{tabular}
\end{center}
\vspace*{-6pt}
\caption{Results of the games between COEVOL* and CRAFTY (W\% is the winning percentage, and RD is the Elo rating difference (see Appendix B)). Win = 1 point, draw = 0.5 point, and loss = 0 point.}
\label{tab:results1}

\end{table}

\vspace*{-6pt}
\subsection{Results of Coevolution}

Repeating the supervised evolutionary process, we obtained each time a ``best evolved organism'' with a different set of evolved parameter values. That is, each run produced a different grandmaster-level program. Even though the  performance of these independently evolved best organisms is fairly similar, our goal was to improve upon these organisms and create an enhanced ``best of best'' organism.

We applied single-population coevolution to enhance the performance of the program. After running the supervised evolution ten times (which ran for about 20 hours), ten different best organisms were obtained. Using these ten organisms as the starting population, we applied GA for 50 generations, where each organism played each other organism four times in every round. Each game was limited to ten seconds (5 seconds per side). In practice, this coevolution phase ran for approximately 20 hours.

We measured the performance of the best evolved organism after coevolution (we call this program \textsc{Coevol*}) by conducting a series of matches against \textsc{Crafty} and also against \textsc{Evol*}. Table~\ref{tab:results2} provides the results of 500 games between \textsc{Coevol*} and \textsc{Evol*}, and between \textsc{Coevol*} and \textsc{Crafty}.

\begin{table}[htbp]

\begin{center}
\begin{tabular}{|l||c|c|c|}
\hline
Match & Result & W\% & RD\\
\hline
\hline
\textsc{Coevol*} - \textsc{Crafty} & 304.5 - 195.5 & 60.9\% & $+77$\\
\hline
\textsc{Coevol*} - \textsc{Evol*}  & 293.0 - 212.0 & 58.6\% & $+60$\\
\hline
\end{tabular}
\end{center}
\vspace*{-6pt}
\caption{Results of the games of COEVOL* against CRAFTY and EVOL*.}
\label{tab:results2}

\end{table}

The results demonstrate that the coevolution phase further improved the performance of the program, resulting in the superiority of \textsc{Coevol*} to both \textsc{Crafty} and \textsc{Evol*}.

\section{Concluding Remarks and \\Future Research}

In this paper we presented a novel approach for evolving grandmaster-level evaluation functions by combining supervised and unsupervised evolution. In contrast to the previous successful attempt which focused on mimicking the evaluation function of a chess program that served as a mentor, the approach presented in this paper focuses on evolving the parameters of interest solely by observing games of human grandmasters, where the only available information to guide the evolution consists of the moves made in these games.

Learning from games of human grandmasters in the supervised phase of the evolution, we obtained several grandmaster-level evaluation functions. Specifically, running the procedure ten times, we obtained ten such evolved evaluation functions, which served as the initial population for the second coevolution phase.

While previous attempts at using coevolution have failed due to the unacceptably large amount of time needed to evolve each generation, the use of coevolution succeeded in our case because the initial population was not random, but relatively well tuned due to the first phase of supervised evolution.

According to our experiments, organisms evolved from randomly initialized chromosomes to sets of highly tuned parameters. The coevolution phase further improved the performance of the program, resulting in an evolved organism which resoundingly defeats a grandmaster-level program. Note that this performance was achieved despite the fact that the evaluation function of the evolved program consists of a considerably smaller number of parameters than that of \textsc{Crafty}, of which the evaluation function consists of over 100 parameters.

In summary, we demonstrated how our approach can be used for automatic tuning of an evaluation function from scratch. Furthermore, the approach can also be applied for enhancing existing highly tuned evaluation functions. Starting from several sets of tuned parameter values of the evaluation function, the coevolution phase can be applied to refine these values, so as to further improve the evaluation function.

Running the supervised evolution phase ten times, together with coevolution, took a total of about 40 hours. Both the supervised and unsupervised phases can be easily parallelized for obtaining linear scalability. During the supervised evolution each organism can be evaluated independently on a different processor, without having to share any information with the other organisms. Also, during coevolution, multiple games can be run in parallel. In this work we ran the experiments on a single processor machine (see Appendix A). Running these tests on an 8-core processor (which is readily available today) would reduce the overall running time from 40 hours to as little as 5 hours.

Finally, the results presented in this paper point to the vast potential in applying evolutionary methods for learning from human experts. We believe that the approach presented in this paper for parameter tuning could be applied to a wide array of problems for essentially ``reverse engineering'' the knowledge of a human expert.

\appendix

\section{Experimental Setup}

\hspace*{-4pt}Our experimental setup consisted of the following resources:

\begin{itemize}
\item \textsc{Crafty 19} chess program running as a native ChessBase engine.

\item \textsc{Fritz 9} interface for automatic running of matches, using \textsc{Shredder} opening book.

\item AMD Athlon 64 3200+ with 1 GB RAM and Windows XP operating system.

\end{itemize}

\section{Elo Rating System}

The Elo rating system, developed by Arpad Elo, is the official system for calculating the relative skill levels of players in chess. The following statistics from the January 2009 FIDE rating list provide a general impression of the meaning of Elo ratings:

\begin{itemize}

\item 21079 players have a rating above 2200 Elo.

\item 2886 players have a rating between 2400 and 2499, most of whom have either the title of International Master (IM) or Grandmaster (GM).

\item 876 players have a rating between 2500 and 2599, most of whom have the title of GM.

\item 188 players have a rating between 2600 and 2699, all of whom have the title of GM.

\item 32 players have a rating above 2700.
\end{itemize}

Only four players have ever had a rating of 2800 or above. A novice player is generally associated with rating values of below 1400 Elo. Given the rating difference ($RD$) of two players, the following formula calculates the expected winning rate ($W$, between 0 and 1) of the player:

\begin{displaymath}
W = \frac{1}{10^{-RD/400} + 1}.
\end{displaymath}

Given the winning rate of a player, as is the case in our experiments, the expected rating difference can be derived from the above formula:

\begin{displaymath}
RD = -400 \log_{10}(\frac{1}{W} - 1).
\end{displaymath}


\begin{thebibliography}{99}

\bibitem{akl77} S.G. Akl and M.M. Newborn.
The principal continuation and the killer heuristic.
In \emph{Proceedings of the Fifth Annual ACM Computer Science Conference}, pages 466--473. ACM Press, Seattle, WA, 1977.

\bibitem{aksenov04} P. Aksenov.
\emph{Genetic algorithms for optimising chess position scoring}.
M.Sc. Thesis, University of Joensuu, Finland, 2004.

\bibitem{anantharaman91} T.S. Anantharaman.
Extension heuristics.
\emph{ICCA Journal}, 14(2):47--65, 1991.

\bibitem{baxter00} J. Baxter, A. Tridgell, L. and Weaver.
Learning to play chess using temporal-differences.
\emph{Machine Learning}, 40(3):243--263, 2000.

\bibitem{beal89} D.F. Beal.
Experiments with the null move.
\emph{Advances in Computer Chess 5}, ed. D.F. Beal, pages 65--79. Elsevier Science, Amsterdam, 1989. 

\bibitem{beal95} D.F. Beal and M.C. Smith.
Quantification of search extension benefits.
\emph{ICCA Journal}, 18(4):205--218, 1995.

\bibitem{bjornsson98} Y. Bjornsson and T.A. Marsland.
Multi-cut pruning in alpha-beta search.
In \emph{Proceedings of the First International Conference on Computers and Games}, pages 15--24, Tsukuba, Japan, 1998.

\bibitem{bjornsson01} Y. Bjornsson and T.A. Marsland.
Multi-cut alpha-beta-pruning in game-tree search.
\emph{Theoretical Computer Science}, 252(1-2):177--196, 2001.

\bibitem{campbell83} M.S. Campbell and T.A. Marsland.
A comparison of minimax tree search algorithms.
\emph{Artificial Intelligence}, 20(4):347--367, 1983.

\bibitem{capablanca06} J.R. Capablanca. \emph{Chess Fundamentals}, ed. N. de Firmian, Random House, Revised ed., 2006.

\bibitem{chinchalkar96} S. Chinchalkar.
An upper bound for the number of reachable positions.
\emph{ICCA Journal}, 19(3):181--183, 1996.

\bibitem{david08a} O. David, M. Koppel, and N.S. Netanyahu.
Genetic algorithms for mentor-assisted evaluation function optimization.
In \emph{Proceedings of the Genetic and Evolutionary Computation Conference}, pages 1469--1476. Atlanta, GA, 2008.

\bibitem{david08b} O. David and N.S. Netanyahu.
Extended null-move reductions.
In \emph{Proceedings of the 2008 International Conference on Computers and Games}, eds. H.J. van den Herik, X. Xu, Z. Ma, and M.H.M. Winands, pages 205--216. Springer (LNCS 5131), Beijing, China, 2008.

\bibitem{donninger93} C. Donninger.
Null move and deep search: Selective search heuristics for obtuse chess programs.
\emph{ICCA Journal}, 16(3):137--143, 1993.

\bibitem{gillogly72} J.J. Gillogly.
The technology chess program.
\emph{Artificial Intelligence}, 3(1-3):145--163, 1972.

\bibitem{gross02} R. Gross, K. Albrecht, W. Kantschik, and W. Banzhaf.
Evolving chess playing programs.
In \emph{Proceedings of the Genetic and Evolutionary Computation Conference}, pages 740--747. New York, NY, 2002.

\bibitem{hauptman05} A. Hauptman and M. Sipper.
Using genetic programming to evolve chess endgame players. 
In \emph{Proceedings of the 2005 European Conference on Genetic Programming}, pages 120--131. Springer, Lausanne, Switzerland, 2005.

\bibitem{hauptman07} A. Hauptman and M. Sipper.
Evolution of an efficient search algorithm for the Mate-in-N problem in chess.
In \emph{Proceedings of the 2007 European Conference on Genetic Programming}, pages 78--89. Springer, Valencia, Spain, 2007.

\bibitem{heinz98a} E.A. Heinz.
Extended futility pruning.
\emph{ICCA Journal}, 21(2):75--83, 1998.

\bibitem{hyatt90} R.M. Hyatt, A.E. Gower, and H.L. Nelson.
\textsc{Cray Blitz}.
\emph{Computers, Chess, and Cognition}, eds. T.A. Marsland and J. Schaeffer, pages 227--237. Springer-Verlag, New York, 1990.

\bibitem{kendall01} G. Kendall and G. Whitwell.
An evolutionary approach for the tuning of a chess evaluation
function using population dynamics.
In \emph{Proceedings of the 2001 Congress on Evolutionary Computation}, pages 995--1002. IEEE Press, World Trade Center, Seoul, Korea, 2001.

\bibitem{nelson85} H.L. Nelson.
Hash tables in \textsc{Cray Blitz}.
\emph{ICCA Journal}, 8(1):3--13, 1985.

\bibitem{reinfeld83} A. Reinfeld.
An improvement to the Scout tree-search algorithm.
\emph{ICCA Journal}, 6(4):4--14, 1983. 

\bibitem{schaeffer83} J. Schaeffer.
The history heuristic.
\emph{ICCA Journal}, 6(3):16--19, 1983. 

\bibitem{schaeffer89} J. Schaeffer.
The history heuristic and alpha-beta search enhancements in practice.
\emph{IEEE Transactions on Pattern Analysis and Machine Intelligence}, 11(11):1203--1212, 1989.  

\bibitem{schaeffer01} J. Schaeffer, M. Hlynka, and V. Jussila.
Temporal difference learning applied to a high-performance game-playing program.
In \emph{Proceedings of the 2001 International Joint Conference on Artificial Intelligence}, pages 529--534. Seattle, WA, 2001.

\bibitem{scott69} J.J. Scott.
A chess-playing program.
\emph{Machine Intelligence 4}, eds. B. Meltzer and D. Michie, pages 255--265. Edinburgh University Press, Edinburgh, 1969.

\bibitem{slate77} D.J. Slate and L.R. Atkin.
\textsc{Chess 4.5} - The Northwestern University chess program.
\emph{Chess Skill in Man and Machine}, ed. P.W. Frey, pages 82--118. Springer-Verlag, New York, 2nd ed., 1983.

\bibitem{sutton98} R.S. Sutton and A.G. Barto.
\emph{Reinforcement Learning: An Introduction}, MIT Press, Cambridge, MA, 1998.

\bibitem{tesauro92} G. Tesauro.
Practical issues in temporal difference learning. 
\emph{Machine Learning}, 8(3-4):257--277, 1992. 

\bibitem{tunstall91} W. Tunstall-Pedoe.
Genetic algorithms optimising evaluation functions.
\emph {ICCA Journal}, 14(3):119--128, 1991.

\bibitem{wiering95} M.A. Wiering.
\emph{TD learning of game evaluation functions with hierarchical neural architectures}.
Master's Thesis, University of Amsterdam, 1995.

\end{thebibliography}
\end{document}